\documentclass[lettersize,journal]{IEEEtran}
\usepackage{amsmath,amsfonts}
\usepackage{algorithmic}
\usepackage{algorithm}
\usepackage{array}
\usepackage[caption=false,font=normalsize,labelfont=sf,textfont=sf]{subfig}
\usepackage{textcomp}
\usepackage{stfloats}
\usepackage{url}
\usepackage{verbatim}
\usepackage{graphicx}
\usepackage{cite}

\usepackage{multirow}
\usepackage[table,xcdraw]{xcolor}
\usepackage{amssymb}
\usepackage{booktabs}
\usepackage{hyperref}

\begin{document}

\title{MambaH-Fit: Rethinking Hyper-surface Fitting-based Point Cloud Normal Estimation via State Space Modelling}

\author{Weijia Wang,~\IEEEmembership{Member,~IEEE}, Yuanzhi Su,~\IEEEmembership{Member,~IEEE}, Pei-Gen Ye,~\IEEEmembership{Member,~IEEE}, Yuan-Gen Wang*,~\IEEEmembership{Senior Member,~IEEE}
\thanks{Weijia Wang and Yuan-Gen Wang are affiliated with Guangzhou University (E-mail: wwj@gzhu.edu.cn, wangyg@gzhu.edu.cn). Yuanzhi Su is affiliated with Hong Kong Polytechnic University (E-mail: yuanzhi.su@connect.polyu.hk). Pei-Gen Ye is affiliated with Beijing Institute of Technology (E-mail: ypgmhxy@bit.edu.cn). *: Corresponding Author.}}

\markboth{Journal of \LaTeX\ Class Files,~Vol.~14, No.~8, August~2021}%
{Shell \MakeLowercase{\textit{et al.}}: A Sample Article Using IEEEtran.cls for IEEE Journals}


\maketitle

\begin{abstract}
We present MambaH-Fit, a state space modelling framework tailored for hyper-surface fitting-based point cloud normal estimation. Existing normal estimation methods often fall short in modelling fine-grained geometric structures, thereby limiting the accuracy of the predicted normals. Recently, state space models (SSMs), particularly Mamba, have demonstrated strong modelling capability by capturing long-range dependencies with linear complexity and inspired adaptations to point cloud processing. However, existing Mamba-based approaches primarily focus on understanding global shape structures, leaving the modelling of local, fine-grained geometric details largely under-explored. To address the issues above, we first introduce an \textit{Attention-driven Hierarchical Feature Fusion (AHFF)} scheme to adaptively fuse multi-scale point cloud patch features, significantly enhancing geometric context learning in local point cloud neighbourhoods. Building upon this, we further propose \textit{Patch-wise State Space Model (PSSM)} that models point cloud patches as implicit hyper-surfaces via state dynamics, enabling effective fine-grained geometric understanding for normal prediction. Extensive experiments on benchmark datasets show that our method outperforms existing ones in terms of accuracy, robustness, and flexibility. Ablation studies further validate the contribution of the proposed components.
\end{abstract}


\section{Introduction}
\IEEEPARstart{E}{stimating} surface normals from 3D point clouds is a fundamental task in many vision-based applications, including point cloud filtering~\cite{Lu_lowrank_2020}, registration~\cite{PPFNet_Deng_2018}, and surface reconstruction~\cite{poisson_2006,screened_poisson_2013}. Raw point clouds lack connectivity information and are often sparse and noisy, making normal estimation a non-trivial problem. For this, the key challenge lies in accurately approximating the surface information from local point cloud neighbourhoods (as patches), from which the surface normals can be derived. Conventional surface fitting methods, such as $n$-jet~\cite{cazals_2005_jet}, approximate explicit polynomial surfaces, which are straightforward and have inspired a set of subsequent learning-based works such as DeepFit~\cite{Shabat_deepfit_2020}, AdaFit~\cite{Zhu_adafit_2021}, and GraphFit~\cite{li_graphfit_2022}. However, $n$-jet fitting requires a pre-defined polynomial order, limiting its flexibility in adapting to varying surface types.

\begin{figure}[t!]
  \centering
  \includegraphics[width=\columnwidth]{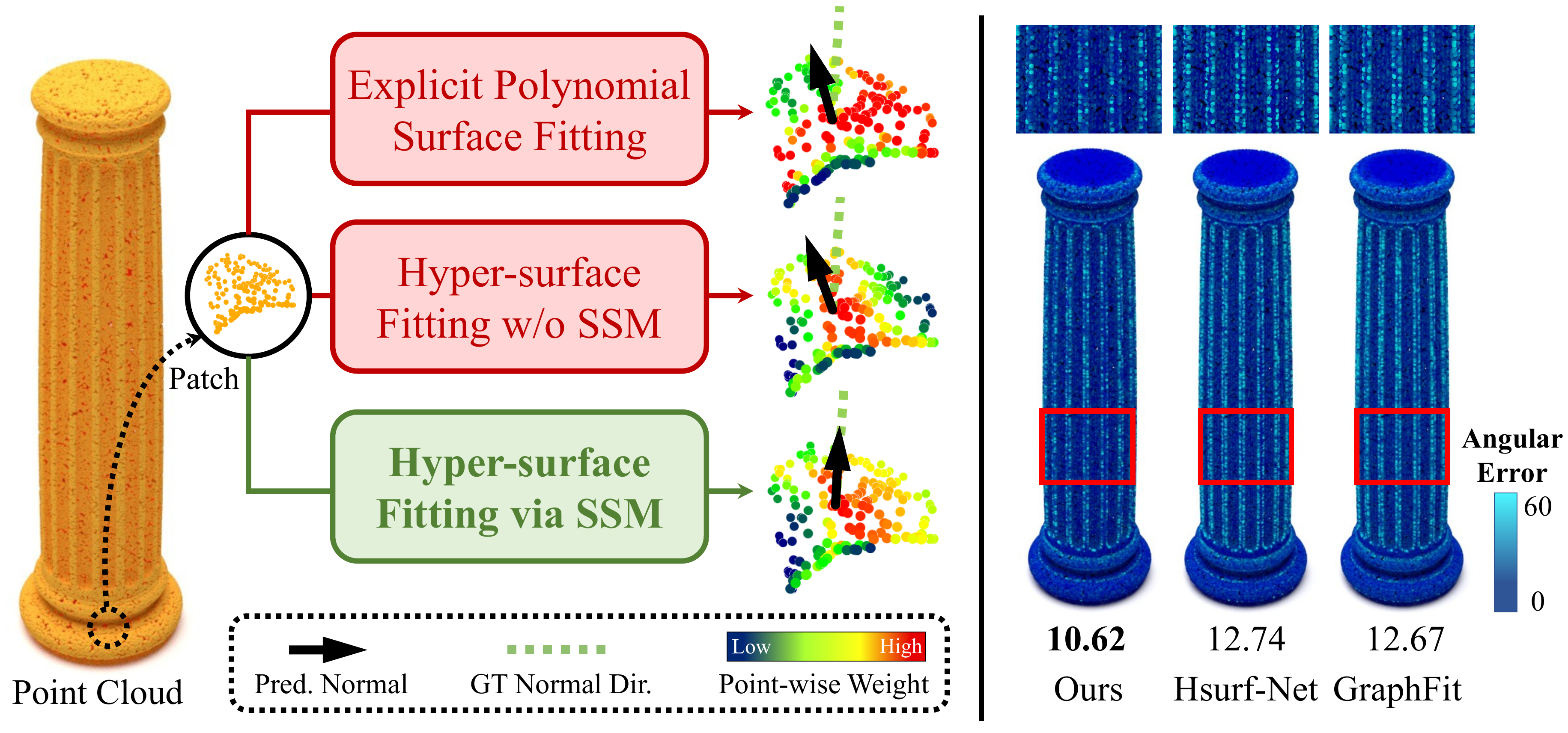}
  \caption{\textit{Left}: Normal estimation results from different surface fitting methods, with comparisons to the ground-truth normal direction and point-wise weights visualised. \textit{Right}: The bottom values indicate RMSE for normal prediction, where our method achieves more accurate results. The angular error is listed in degrees. }
  \label{fig:teaser}
\end{figure}

To alleviate the issue above, HSurf-Net~\cite{li_2022_hsurf} replaces the explicit surface fitting process with operations in a latent space. Specifically, it uses residual blocks to learn implicit hyper-surface representations for input point clouds. While this approach eliminates the need for a pre-defined polynomial order, its residual block structure processes point-wise features independently, without explicitly modelling inter-point relationships within the patch. This limits the network's ability to learn coherent geometric patterns on complex surfaces, especially in noisy or sparse settings.
Due to the discrete and unordered nature of point clouds, modelling latent geometric relationships within the feature space is crucial for learning accurate surface information
~\cite{wang_dgcnn_2019,pistilli_2020_gpdnet,Mao_2024_PDLTS}. To effectively model such relationships, a strong self-learning mechanism is required, where Transformer~\cite{vaswani_transformer_2017} offers a competitive solution by implicitly learning inter-point relationships via a self-attention mechanism. Nonetheless, its quadratic complexity in memory and computation makes it unsuitable for dense, large-scale point clouds, and its performance on learning fine-grained geometric features remains limited.

Recently, state space models (SSMs), particularly Mamba~\cite{gu_mamba_colm_2024}, have gained increasing attention for their ability to efficiently model long-range dependencies. Compared with Transformer-based structures that scale quadratically with respect to input sequence length, Mamba enables efficient linear-time modelling and demonstrates promising performance across various domains, including natural language processing~\cite{gu_mamba_colm_2024}, 2D computer vision~\cite{kui_glmambanet_2025}, and 3D point cloud analysis~\cite{liang_pointmamba_24,zhang_pcm_25}. Existing Mamba-based architectures primarily focus on capturing the global shape information of point clouds, benefiting tasks such as classification and segmentation. However, they overlook the importance of fine-grained geometric details, including surface variations and curvature changes, which are crucial for accurate normal prediction. To capture fine geometric details, existing normal estimation methods commonly adopt hierarchical patch feature fusion strategies. These approaches typically integrate features from larger point neighbourhoods into smaller ones. Nevertheless, such fusion schemes typically perform simple max-pooling on point features of larger scales, lacking flexibility to emphasise relevant regions and suppress irrelevant ones.

Motivated by these observations, we rethink the hyper-surface fitting process and propose \textit{MambaH-Fit}, a novel framework tailored for point cloud normal estimation. As illustrated in Fig.~\ref{fig:teaser}, our method captures fine-grained geometric details and achieves more accurate results compared with existing methods. MambaH-Fit includes an \textit{Attention-driven Hierarchical Feature Fusion (AHFF)} module, enabling the network to flexibly learn relevant and irrelevant geometric features, thereby facilitating accurate learning of geometric details. It also includes a \textit{Patch-wise State Space Model (PSSM)} that takes point features as input token sequences and models implicit surfaces using lightweight Mamba blocks, facilitating accurate tangent plane approximation and subsequent normal derivation on point cloud patches. Extensive experiments show that our proposed MambaH-Fit achieves competitive performance across a diverse range of point cloud surface types, especially on large-scale real-world scanned datasets.

In summary, our contributions are threefold:
\begin{itemize}
    \item We present MambaH-Fit, the first Mamba-based framework for accurate point cloud normal estimation, where experimental results demonstrate competitive performance in terms of accuracy, robustness, and efficiency. 
    \item We propose AHFF, an attention-driven hierarchical scheme that adaptively extracts global features of larger point cloud neighbourhoods and fuses them into smaller ones, facilitating accurate normal estimation results.
    \item We design PSSM, a patch-wise SSM tailored for hyper-surface fitting that enables effective local geometry modelling via Mamba. It demonstrates superior capability in tangent plane approximation and subsequent normal estimation. 
\end{itemize}

\section{Related Work}

\subsection{Conventional Normal Estimation Methods}

The Principal Component Analysis (PCA) normal estimation method~\cite{Hoppe_1992} computes the normal of each point via plane fitting on local neighbourhoods. While straightforward, this method is sensitive to the chosen neighbourhood scale and often delivers rough normals that fail to effectively preserve the underlying surface features.
To better retain local geometric features, $n$-jet fitting~\cite{cazals_2005_jet} was introduced, which fits a polynomial surface onto the input point cloud and derives the normal vector subsequently. Compared with PCA, $n$-jet fitting can model complex local structures more precisely.
Other conventional normal estimation methods include Voronoi-based algorithms~\cite{Amenta_1998,DEY_2006,Alliez_2007,Merigot_2010},
randomised Hough transform~\cite{Boulch_2012},
edge-aware resampling~\cite{EAR2013},
and low-rank matrix approximation~\cite{Lu_lowrank_2020}.
Although these methods demonstrate improved capability for feature preservation, they typically require non-trivial parameter tuning to achieve desirable normal estimation results. Besides, they exhibit limited robustness when applied to point clouds that are contaminated with noise or come with complex geometric structures.

\subsection{Learning-based Normal Estimation Methods}
With the advancements in deep learning-driven point cloud analysis, learning-based methods have demonstrated stronger generalisation capabilities, eliminating the need for cumbersome parameter tuning required by conventional methods.
PCPNet~\cite{Guerrero_pcpnet_2018}, for instance, leverages PointNet's~\cite{Qi_2017_pointnet} feature extraction modules, providing a pioneering learning-based benchmark for normal estimation.
Nesti-Net~\cite{Benshabat_Nesti_2019} introduces a mixture-of-experts network to enhance the robustness of normal estimation, enabling the network to flexibly learn different types of point cloud surfaces. However, directly regressing normals makes the network less adaptive to complex surface types, which has motivated the emergence of methods incorporating surface-fitting concepts.
For example, IterNet~\cite{Lenssen_DI_2020} employs a deep graph-based network that refines normals by iteratively parameterising an adaptive kernel and performing weighted least-squares plane fitting.
Inspired by the $n$-jet fitting, DeepFit~\cite{Shabat_deepfit_2020}, AdaFit~\cite{Zhu_adafit_2021}, and GraphFit~\cite{li_graphfit_2022} were proposed as learning-based solutions, which predict point-wise weights to approximate explicit polynomial surfaces onto local point cloud neighbourhoods and further derive normals. Based on these methods, ZTEE~\cite{Du_2023_CVPR} additionally considers the $z$-direction transform to achieve better surface fitting results. However, such methods require a fixed jet order as the training hyper-parameter, limiting the network's adaptability to a wider range of surface types during testing.
To overcome this limitation, HSurf-Net~\cite{li_2022_hsurf} employs high-dimensional feature codes and residual blocks to learn implicit surface representations, showing improved adaptiveness in varying surface geometries.
SHS-Net~\cite{li2023shsnet} and NGLO~\cite{li_nglo_23} further advance implicit surface fitting by introducing Signed Distance Functions (SDF) and neural gradient learning strategies. 
NeAF~\cite{li2023neaf} learns an angle field around each query normal and gradually refines the prediction's accuracy.
To achieve more robust normal estimation performance on point clouds with varying noise levels, CMG-Net~\cite{wu_cmgnet_24} incorporates a patch fusion strategy and designs an alternative training loss to improve the robustness of the feature encoding network. 
More recently, HAE~\cite{Li_HAE_2025_CVPR} combines angular and Euclidean distance encoding as a hybrid technique, demonstrating improved accuracy and robustness in normal prediction. In addition, an emerging line of research has explored joint methods that estimate normals while simultaneously reducing noise from point clouds, such as PCDNF~\cite{zheng_pcdnf_2024}, GeoDualCNN~\cite{wei_geodualcnn_2023}, and the method based on contrastive learning~\cite{Edi_contrast_2024}. While prior works have made notable progress in the field of point cloud normal estimation, they largely overlook the modelling of long-range spatial dependencies that are crucial for learning complex geometric features. This motivates our exploration into SSMs for context-aware modelling on fine-grained local surface details.

\subsection{State Space Models in 3D Vision} 
State space models have long been foundational in control domains for modelling dynamical systems. Motivated by their sequential modelling capabilities, Mamba~\cite{gu_mamba_colm_2024} employs a selective SSM mechanism that enables efficient training and inference with linear computational complexity. In contrast to the quadratic complexity of Transformers, Mamba leverages hardware-aware optimisation and selective state updates, making it a competitive solution for modelling long sequences~\cite{wang_mamba_24,zhao_cobra_25}. Building on these strengths, recent works have adapted SSMs to 3D vision tasks. For instance, Mamba3D~\cite{xu_mamba3d_24} replaces self-attention with directional SSM blocks to capture long-range dependencies in point clouds. PointMamba~\cite{liang_pointmamba_24} and PCM~\cite{zhang_pcm_25} employ serialisation techniques that enable global point cloud modelling through SSM propagation. 
Voxel Mamba~\cite{zhang_voxel_24} serialises unordered voxel features into structured sequences via Hilbert curves, enabling group-free SSM-based processing for 3D object detection. Similarly, Grid Mamba~\cite{yang_grid_25} unifies five space-filling curves under a single encoding theory and proposes Grid Multiview Scanning to mitigate spatial proximity loss during voxel serialisation. To improve modelling flexibility, PointRamba~\cite{wang_pointramba_24} combines Mamba’s efficiency with Transformer’s dynamic attention for point cloud analysis. Beyond point cloud global feature learning, 3DMambaIPF~\cite{zhou_3dmambaipf_25} demonstrates Mamba's capability for iteratively refining geometric details through stacked SSM modules. Motivated by Mamba’s capability in modelling sequential data, we develop a new state-space modelling framework that fits implicit hyper-surfaces over local point cloud neighbourhoods, effectively capturing fine-grained geometry information for accurate normal estimation.

\section{Proposed Method}

\subsection{Normal Estimation Preliminaries}
Given a 3D point cloud $\mathbf{P} = \{\mathbf{p}_1,\ldots, \mathbf{p}_T\ |\ \mathbf{p}_i \in \mathbb{R}^3,\ i\ = 1, ..., T \}$ with $T$ points in total, the goal is to predict the \textit{normal} of each point $\mathbf{p}$, which is perpendicular to the tangent plane at that point. To achieve this, we construct a local point cloud neighbourhood $\mathcal{P}$ (i.e., a \textit{patch}) centered at the query point $\mathbf{p}$, consisting of its $N-1$ nearest neighbours, which collectively represent the local geometric surface. 

\begin{figure*}[t]
  \centering
  \includegraphics[width=\textwidth]{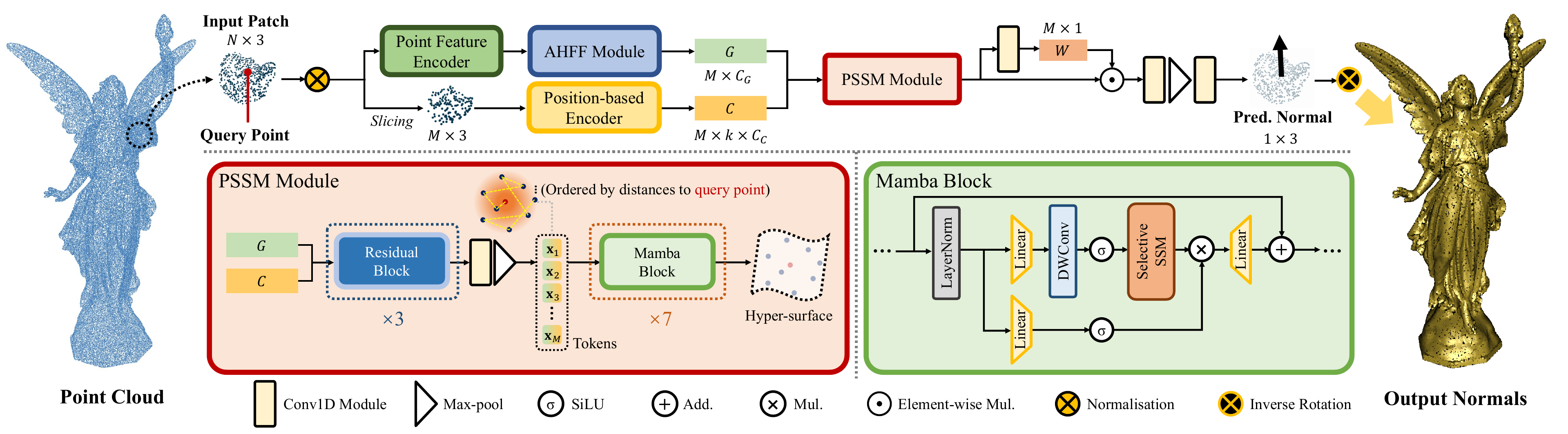}
  \caption{Overview of MambaH-Fit's architecture. Our AHFF module enables adaptive hierarchical feature fusion for $G$. $C_G$ and $C_C$ represent encoding dimensions for feature codes $G$ and $C$, respectively. In our PSSM module, point features $\{\mathbf{x}_1, \ldots, \mathbf{x}_M\}$ (as tokens) are naturally ordered by each point's spatial distance to the patch's query point. }
  \label{fig:network}
\end{figure*}

It is noteworthy that the raw patch $\mathcal{P}$ may be of any size and contain arbitrary rigid transformations (e.g., rotations and translations), making it infeasible for deep neural networks to effectively extract the geometric features when directly used as input. To solve this issue during training, we first normalise $\mathcal{P}$ into a unit sphere by dividing its radius, and align it with the Cartesian coordinate system using a rotation matrix $\mathbf{R}$ that is computed via PCA decomposition on the patch $\mathcal{P}$. The patch is then input into encoding modules for geometric feature extraction, from which the normal is subsequently regressed. During testing phase, we normalise the predicted normal's length and map it back to its original orientation using the inverse rotation matrix $\mathbf{R}^{-1}$. Our overall network architecture is illustrated in Fig.~\ref{fig:network}.

\subsection{Revisiting Hyper-surface Fitting}

Given a point cloud patch $\mathcal{P} = \{\mathbf{p}_i = (x_i, y_i, z_i)\ |\ i\ = 1, ..., N \}$, the conventional $n$-jet fitting method~\cite{cazals_2005_jet} aims to fit a polynomial surface onto the points. It assumes the $x$-axis and $y$-axis form the tangent space, and expresses the surface height $z$ as a function of $x$ and $y$, i.e., $z = f(x, y)$. For any point $\mathbf{p}_i$ within $\mathcal{P}$, the function value $f(x_i, y_i)$ is approximated by the truncated Taylor expansion $J_{\alpha, n}(x_i, y_i)$, where $\alpha$ denotes the vector of Taylor coefficients up to order $n$. To ensure the optimal $\alpha$ is found, the least squares approximation method is applied to minimise the difference between $J_{\alpha, n}(x_i, y_i)$ and the height function $f(x_i, y_i)$, and the optimal solution $J_{\alpha, n}^*$ is written as
\begin{equation}
\label{eq:height-func}
  J_{\alpha, n}^* = \mathop{\textup{argmin}}\limits_{\alpha}\sum_{i=1}^{N}(f(x_i,y_i)-J_{\alpha, n}(x_i, y_i))^2.
\end{equation}
Using the optimal $\alpha$, the estimated normal vector $\hat{\mathbf{n}}_\mathbf{p}$ is computed by
\begin{equation}
\label{eq:nml-h-alpha}
  \hat{\mathbf{n}}_\mathbf{p} = h(\alpha) = \frac{(-\alpha_{1,0},-\alpha_{0,1}, 1)}{\sqrt{1 + \alpha_{1,0}^2 + \alpha_{0,1}^2}}.
\end{equation}

Since $n$-jet fitting requires a pre-defined polynomial order $n$, its flexibility in modelling various surfaces remains limited. To address this limitation, HSurf-Net~\cite{li_2022_hsurf} leverages neural networks to learn implicit hyper-surfaces adaptively for input point clouds. Specifically, the input 3D coordinates are reformulated into high-dimensional features $(G, C, F)$, where $G$ and $C$ denote high-dimensional point cloud features, and $F$ represents the target implicit surface value to be approximated. The surface value $F$ is modelled as a height function $\mathcal{F}$ over the $c$-dimension feature space, such that $F \sim \mathcal{F}(G, C), F \in \mathbb{R}^c$. To approximate this unknown function, a bivariate function $\mathcal{N}_{\theta, \tau}(G, C)$ is introduced, which is implemented using a set of residual blocks with parameters $\theta$ and $\tau$. By replacing the terms in Eq.~\eqref{eq:height-func}, the hyper-surface fitting objective becomes
\begin{equation}
\mathcal{N}_{\theta, \tau}^{*}=\underset{\theta, \tau}{\text{argmin}} \sum_{i=1}^{N}\left\| \mathcal{N}_{\theta, \tau}\left(G_{i}, C_{i}\right)-\mathcal{F}\left(G_{i}, C_{i}\right)\right\| ^{2},
\label{eq:hsurf-fit}
\end{equation}
where $\mathcal{N}_{\theta, \tau}^{*}$ is the optimal hyper-surface.
The feature codes $G$ and $C$ are extracted by different modules:
$G$ contains hierarchical geometric features encoded by Dense Blocks~\cite{Huang_denseblock_17}, while $C$ consists of relative positional information of each point and its local $k$ neighbours, respectively.
Similar to Eq.~\eqref{eq:nml-h-alpha}, the normal is regressed via a mapping function $\mathcal{H}:\mathbb{R}^c \rightarrow \mathbb{R}^3$ as
\begin{equation}
    \mathbf{\hat{n}}_{p}=\mathcal{H}\left(\text{MAX}\left\{w_{i} \cdot \mathcal{N}_{\theta, \tau}\left(G_{i}, C_{i}\right) | i=1, ..., N\right\}\right),
\label{eq:normal-est}
\end{equation}
where $\text{MAX}$ denotes max-pooling, and $w_i$ represents point-wise weight that is regressed from $\mathcal{N}_{\theta, \tau}\left(G_{i}, C_{i}\right)$ and is leveraged for normal prediction. 

Eqs.~\eqref{eq:hsurf-fit} and~\eqref{eq:normal-est} demonstrate that the quality of the implicit surface $\mathcal{N}_{\theta, \tau}\left(G, C\right)$ is critical to accurate normal estimation. However, the fitting scheme in HSurf-Net suffers from two key limitations that hinder the learning of high-quality hyper-surfaces. 
First, the hierarchical feature fusion scheme of $G$ relies on simple max-pooling without the capacity to distinguish relevant features from irrelevant ones, limiting the network's ability to adaptively incorporate geometric context during multi-scale patch feature fusion.
Second, each point feature is processed independently by residual blocks during the fitting process, while the latent intra-patch geometric relationships among the points are under-explored. Without explicitly learning informative geometric cues from distant yet correlated points, this approach struggles to accurately fit tangent planes and estimate normals in the subsequent steps, which is especially evident on complex surfaces or irregular point distributions.

In response to the aforementioned challenges, we propose two novel schemes for high-quality hyper-surface fitting: (1) we design the AHFF module that exploits an attention-guided scheme for hierarchical, multi-scale patch feature fusion; (2) we introduce PSSM that is specifically designed for fitting implicit hyper-surfaces on patches via Mamba blocks, significantly enhancing the subsequent normal estimation results. The proposed two modules will be detailed in subsections~\ref{sec:ahff} and~\ref{sec:pssm}, respectively.


\begin{figure}[t]
  \centering
  \includegraphics[width=\columnwidth]{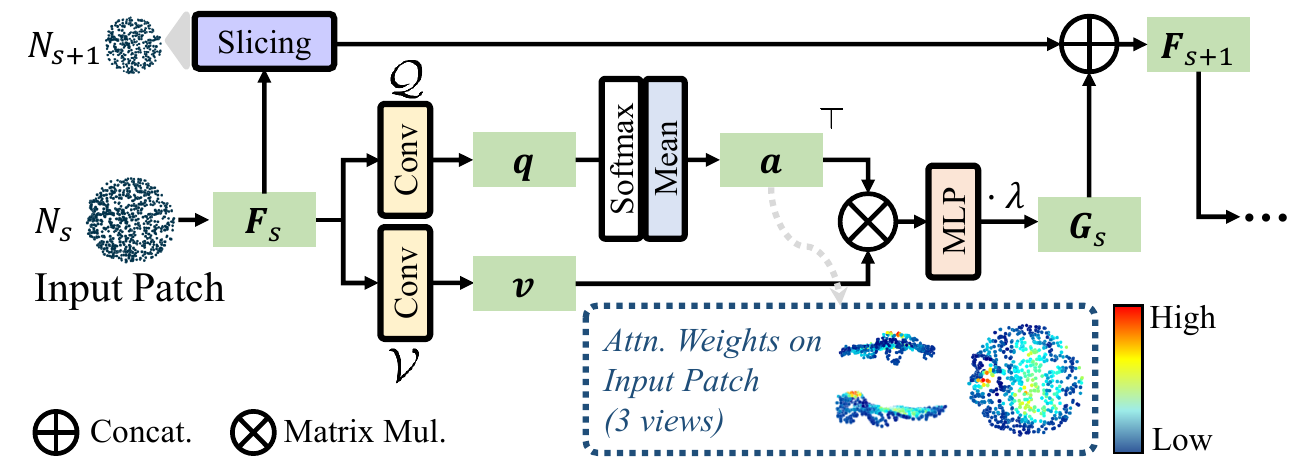}
  \caption{Illustration of our AHFF module's attention-driven fusion scheme. We visualise the attention weights on the $N_s$-point patch, where geometric features (e.g., fluctuating regions), which are not necessarily near the patch centre, are assigned higher weights.}
  \label{fig:ahff-module}
\end{figure}

\subsection{Attention-guided Hierarchical Feature Fusion}
\label{sec:ahff}
Recent works~\cite{Zhu_adafit_2021,li_2022_hsurf,li2023shsnet,wu_cmgnet_24} incorporate hierarchical fusion of patch features across different neighbourhood scales (e.g., $N_s$-point and $N_{s+1}$-point patches, where $N_{s+1} \leq N_{s}$), aiming to enrich geometric representations by combining multi-scale contextual information. A basic fusion technique can be summarised as
\begin{equation}
    \mathbf{F}_{s+1} = \Phi \left(\text{MAX}\left\{\mathbf{F}_{s, j} \right\},\mathbf{F}_{s,i}\right),
\end{equation}  
where $i=1,\ldots,N_{s+1}$ and $j=1,\ldots,N_s$, 
$\Phi$ represents a set of MLP modules, $\mathbf{F}_s$ and $\mathbf{F}_{s+1}$ represent the features of the input and output patch, respectively.
Nonetheless, this aggregation technique simply obtains the global feature vector via max-pooling, treating each point within scale $N_{s+1}$ equally. As a result, the scheme does not adaptively embed the geometric contexts into $\mathbf{F}_{s+1}$ and may overlook subtle yet informative geometric clues. 

To overcome this limitation, inspired by the attention-weighted normal prediction module in SHS-Net~\cite{li2023shsnet}, our proposed AHFF module adaptively selects patch features and embeds them into the global feature vector via a modified attention mechanism. We first obtain two feature matrices $\mathbf{q}$ and $\mathbf{v}$ from $\mathbf{F}_{s}$ via convolutional modules $\mathcal{Q}$ and $\mathcal{V}$, such that $\mathbf{q} = \mathcal{Q(\mathbf{F_{s}})}$ and $\mathbf{v} = \mathcal{V(\mathbf{F_{s}})}$.
Then, the attention score $\mathbf{a}$ is computed by
\begin{equation}
\label{eq:ahff-mean}
    \mathbf{a} = \frac{1}{D} \sum_{d=1}^{D} \text{Softmax}_{N_s}(\mathbf{q}_d),
\end{equation}
where $\mathbf{q}$ is normalised along the point dimension $N_s$ via $\text{Softmax}$, followed by computing the mean value over the encoding dimension $D$. The weighted global feature vector $\mathbf{G}_{s}$ is then computed by
\begin{equation}
\label{eq:attn-global}
    \mathbf{G}_{s} = \Psi\left( \mathbf{v} \cdot \mathbf{a}^\top \right) \cdot \lambda,
\end{equation}
where $\top$ denotes transpose, $\lambda$ is a learnable scale and is initially set to 0.1, $\Psi$ represents MLP modules. Finally, the output feature matrix of the $N_{s+1}$-point patch is computed by
\begin{equation}
    \mathbf{F}_{s+1, i} = \Theta\left( \mathbf{G}_{s}, \, \mathbf{F}_{s, i} \right), \quad i = 1, \dots, N_{s+1},
\end{equation}
where $\Theta$ represents a convolutional module that is applied after concatenating the duplicated $\mathbf{G}_{s}$ and sliced $\mathbf{F}_{s}$.


The fusion scheme of our AHFF module is illustrated in Fig.~\ref{fig:ahff-module}. We adopt the same hierarchical patch scales as HSurf-Net for fusion, where the final fused representation $G$ contains $M=N/4$ point features. Our attention-driven scheme enables the network to implicitly learn the relevance of geometric regions, and its effectiveness will be further validated by the ablation results in Table~\ref{tab:ablation}.

\subsection{Patch-wise State Space Model}
\label{sec:pssm}




An SSM can effectively model structured sequences by mapping the input $x_t$ to output $y_t$ through a latent state $h_t$ that evolves over time $t$. To adapt to discrete point clouds, the system first needs to be discretised~\cite{gu_mamba_colm_2024} such that
\begin{equation} 
\label{eq:discrete-ssm}
    h_t = \bar{\mathbf{A}} h_{t-1} + \bar{\mathbf{B}} x_t, \quad  y_t = \mathbf{C} h_{t},
\end{equation}
where $\bar{\mathbf{A}}$, $\bar{\mathbf{B}}$ and $\mathbf{C}$ are parameters of the system.
To enable efficient computation on GPUs, Eq.~\eqref{eq:discrete-ssm} is reformulated via global convolution as
\begin{equation} 
    \bar{\mathbf{K}} = (\mathbf{C}\bar{\mathbf{B}}, \mathbf{C}\bar{\mathbf{A}}\bar{\mathbf{B}}, \ldots, \mathbf{C}\bar{\mathbf{A}}^{L-1}\bar{\mathbf{B}}), y = x * \bar{\mathbf{K}},
\end{equation}
where $L$ represents the input sequence's length, and $\bar{\mathbf{K}}$ is the kernel of the global convolution.

This formulation enables Mamba to incorporate global context across the input sequence with linear computational complexity, thereby demonstrating strong modelling capability in long-range dependencies. We argue that this mechanism can be effectively adapted to point cloud patches, where modelling intra-patch geometric correlations among the points is crucial. Based on this insight, we design PSSM, a patch-level SSM that is specifically tailored for implicit surface fitting in the latent space.
As Mamba requires sequential input to capture long-range dependencies, we first build point cloud tokens that depict the surface information of the input patch. Following HSurf-Net's technique, we encode the patch into serialised tokens $\mathbf{T}$ by
\begin{equation}
    \mathbf{T} = \Xi(G, C) = \text{MAX}_k\left\{ \mathcal{C}\left( \mathcal{R} \left[ G : C \right] \right) \right\},
\end{equation}
where $\Xi$ is a composite module consisting of a residual MLP stack $\mathcal{R}$ applied to the concatenated input $[G : C]$, followed by an output convolutional module $\mathcal{C}$ and a max-pooling operation over the $k$-local neighbour dimension for each point. As each patch is constructed by identifying the query point $\mathbf{p}$'s $N-1$ nearest neighbours, the resulting tokens $\mathbf{T}$ is natually ordered by proximity to $\mathbf{p}$. As illustrated in PSSM's structure in Fig.~\ref{fig:network}, this ordering implicitly encodes spatial context and aligns well with the sequential modelling paradigm of Mamba.

\begin{figure}[t]
  \centering
  \includegraphics[width=\columnwidth]{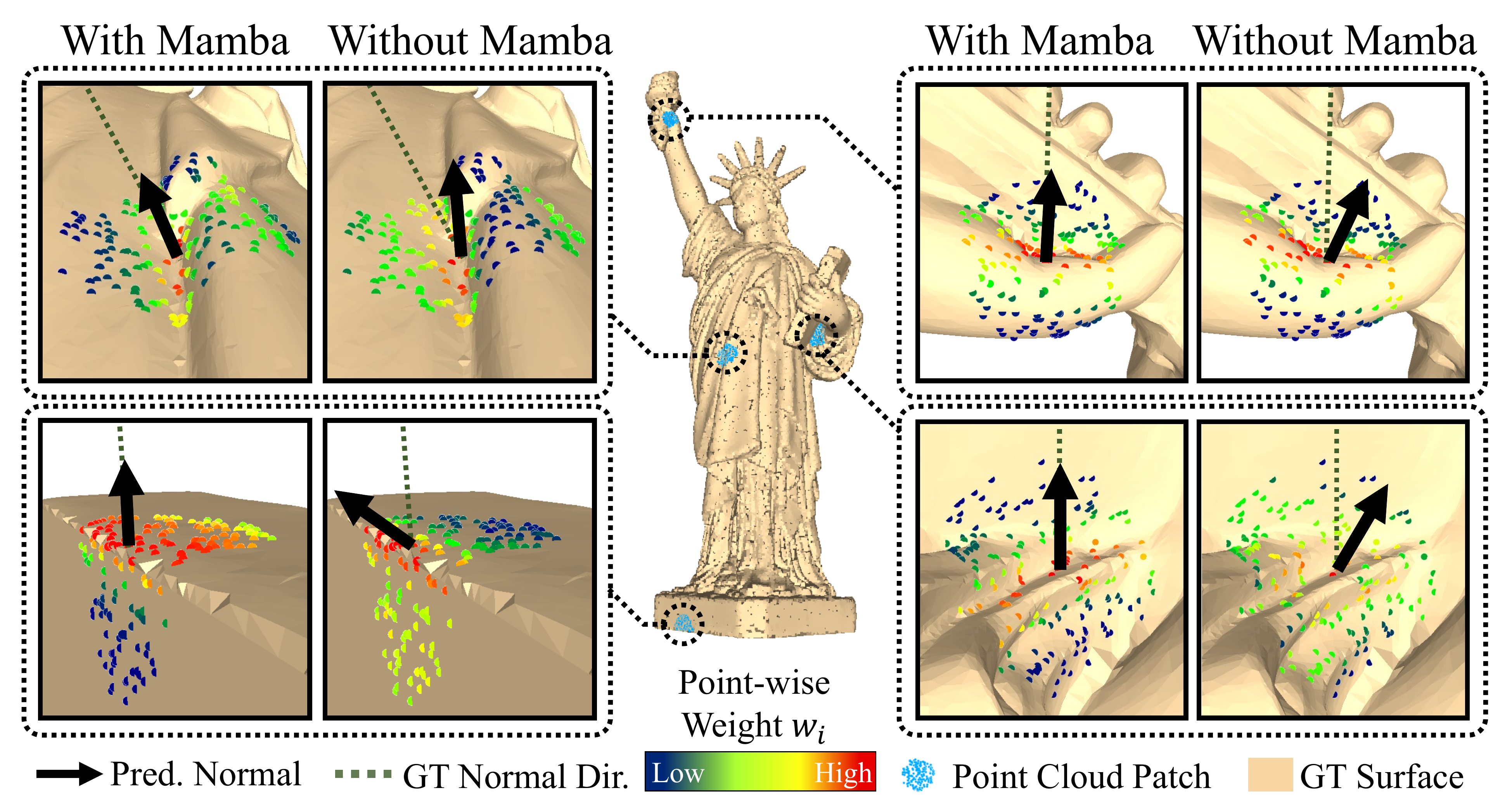}
  \caption{Demonstration of Mamba's effectiveness on diverse surface geometries. We visualise the predicted normal's accuracy, point-wise weight $w_i$, and each patch's corresponding mesh surface.}
  \label{fig:with-mamba-demo}
\end{figure}

The tokens are then input into a chained Mamba module $\mathcal{M}$, which selectively propagates or suppresses information across the sequence via state-space transitions, and outputs the hyper-surface eventually. We denote this process as
\begin{equation}
    \mathcal{N}_{\theta, \tau}\left(G, C\right) = \mathcal{M}(\mathbf{T})= \mathcal{M}(\Xi(G, C)),
\end{equation}
where 7 Mamba blocks are used. As shown in Fig.~\ref{fig:network}, the transition process of the $l$-th Mamba block is defined as
\begin{gather}
    \mathbf{T}_{l}'=\text{DWConv}\left(\text{Linear}\left(\text{LN}\left(\mathbf{T}_{l-1}\right)\right)\right), \notag \\
    \mathbf{s}_{l}=\text{Linear}\left(\text{SSM}\left(\sigma\left(\mathbf{T}_{l}'\right)\right) \times \sigma\left(\text{Linear}\left(\text{LN}\left(\mathbf{T}_{l-1}\right)\right)\right)\right), \notag \\
    \mathbf{T}_{l}=\mathbf{s}_{l}+\mathbf{T}_{l-1},
\end{gather}
where $\text{DWConv}$ represents depth-wise convolution, $\text{LN}$ denotes layer normalisation, and $\sigma$ is SiLU activation function. This recursive process significantly enhances the quality of hyper-surface learning, capturing intra-patch relationships among the points within the patch. A high-quality hyper-surface leads to more accurate tangent plane approximation at query point $\mathbf{p}$, which determines the precision of the estimated normal. 
As illustrated in Fig.~\ref{fig:with-mamba-demo}, Mamba enhances the context-aware encoding ability of the network, showing effectiveness in point-wise weight prediction and normal estimation over various types of surfaces. 

\subsection{Normal Output Module}

The hyper-surface is high-dimensional, and estimating point-wise weights can approximate the tangent plane for normal derivation. To realise this, following the scheme initially proposed in~\cite{zhang_2022_geometryguide}, we first estimate the point-wise weight $ w_i$ by
\begin{equation} 
    w_i = c + \text{sigmoid}(\phi(\mathcal{N}_{\theta, \tau}\left(G_{i}, C_{i}\right))), i=1, \ldots, M,
\end{equation} 
where $\phi$ represents an MLP module, and $c$ is a small constant value. Then, based on Eq.~\eqref{eq:normal-est}, the weights are utilised to predict normal $\hat{\mathbf{n}}_p$ via a mapping function $\mathcal{H}$, which is implemented using an MLP module. $\hat{\mathbf{n}}_p$ is a $1 \times 3$ vector and is normalised to unit length.

\subsection{Loss Functions}

We first minimise the sine loss $\mathcal{L}_\text{sin}$ between the predicted normal $\hat{\mathbf{n}}_{p}$ and the ground-truth $\mathbf{n}_{p}$ by defining
\begin{equation} 
    \mathcal{L}_\text{sin} = \left\| \hat{\mathbf{n}}_{p} \times \mathbf{n}_{p} \right\|.
\end{equation} 
Additionally, to constrain the spatial relevance among the input data points, we adopt the weight loss $\mathcal{L}_\text{wt}$ following~\cite{zhang_2022_geometryguide} to facilitate tangent plane approximation by defining
\begin{equation}
    \mathcal{L}_\text{wt} = \frac{1}{M}\sum_{m=1}^{M}(\hat{w}_{m} - {w}_{m})^2,
\end{equation}
where $\hat{w}_m = \exp (-{(|\mathbf{p}_{m} \cdot \mathbf{n}_{p}|)}^2 / \delta)$ is the predicted point-wise weight, $\delta = \max \left(0.0025,\ 0.3 * (1/M \sum_{m=1}^{M} {d_{m}}^2) \right)$, $d_m = |\mathbf{p}_{m} \cdot \mathbf{n}_{p}|$. Finally, the total loss $\mathcal{L}$ is formulated as
\begin{equation}
    \mathcal{L} = \gamma_1 \mathcal{L}_\text{sin} + \gamma_2 \mathcal{L}_\text{wt}
\end{equation}
where $\gamma_1$ and $\gamma_2$ are weighting factors that are empirically set to 0.1 and 1.0, respectively. 


\section{Experimental Results}

\subsection{Implementation Details}
For fair comparison, we follow the same training settings as in PCPNet~\cite{Guerrero_pcpnet_2018}, including the same train-test split and data augmentation techniques. Our network model is implemented using PyTorch 1.11.0 and is trained on an NVIDIA GeForce RTX 4090 GPU. We set patch size $N$ to 700, train the network for 900 epochs in a batch size of 100 using a single Adam optimiser. The initial learning rate is 0.0005 and is multiplied by a factor of 0.2 in the 200$^{th}$, 400$^{th}$, 600$^{th}$ and 800$^{th}$ epoch, respectively. Following prior works, we train our network model on PCPNet's training set and evaluate its generalisation ability on other benchmarks without fine-tuning.

\subsection{Evaluation Metrics}
We adopt Root Mean Squared Error (RMSE), a straightforward metric that measures point-wise error compared with ground-truth normals, where lower values indicate more accurate results. Additionally, we evaluate the Proportion of Good Points $\alpha$ (PGP$\alpha$) metric (also known as Area Under the Curve, AUC), which measures the proportion of estimated normals under error thresholds of $\alpha$ degrees. The $\alpha$ value is set to 5, 10, 15, 20, 25 and 30, respectively.

\subsection{Results on Synthetic Benchmark}
\begin{figure}[!t]
  \centering
  \includegraphics[width=\columnwidth]{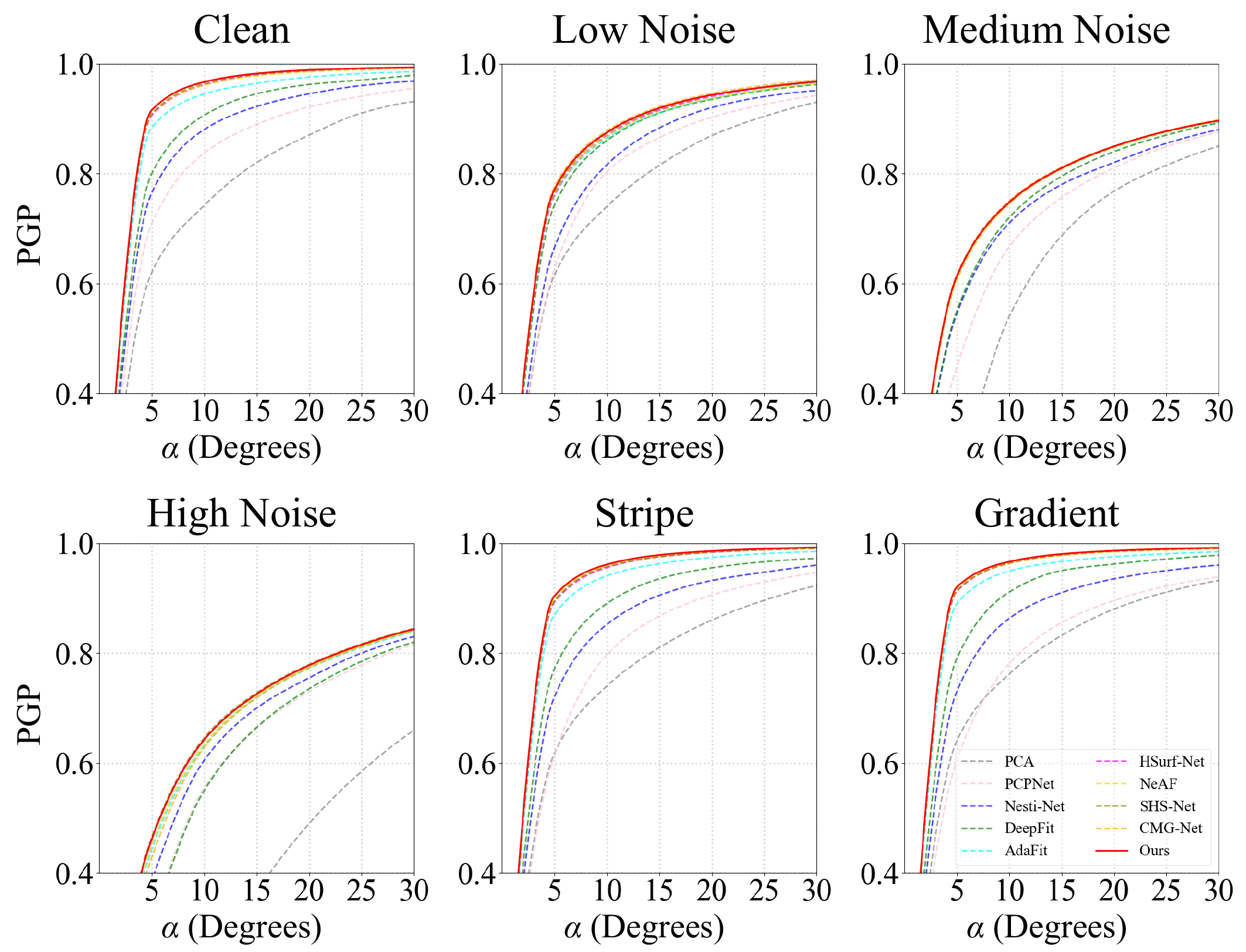}
  \caption{PGP$\alpha$ on different PCPNet test settings.}
  \label{fig:result-auc}
\end{figure}

\begin{figure*}[t]
  \centering
  \includegraphics[width=\textwidth]{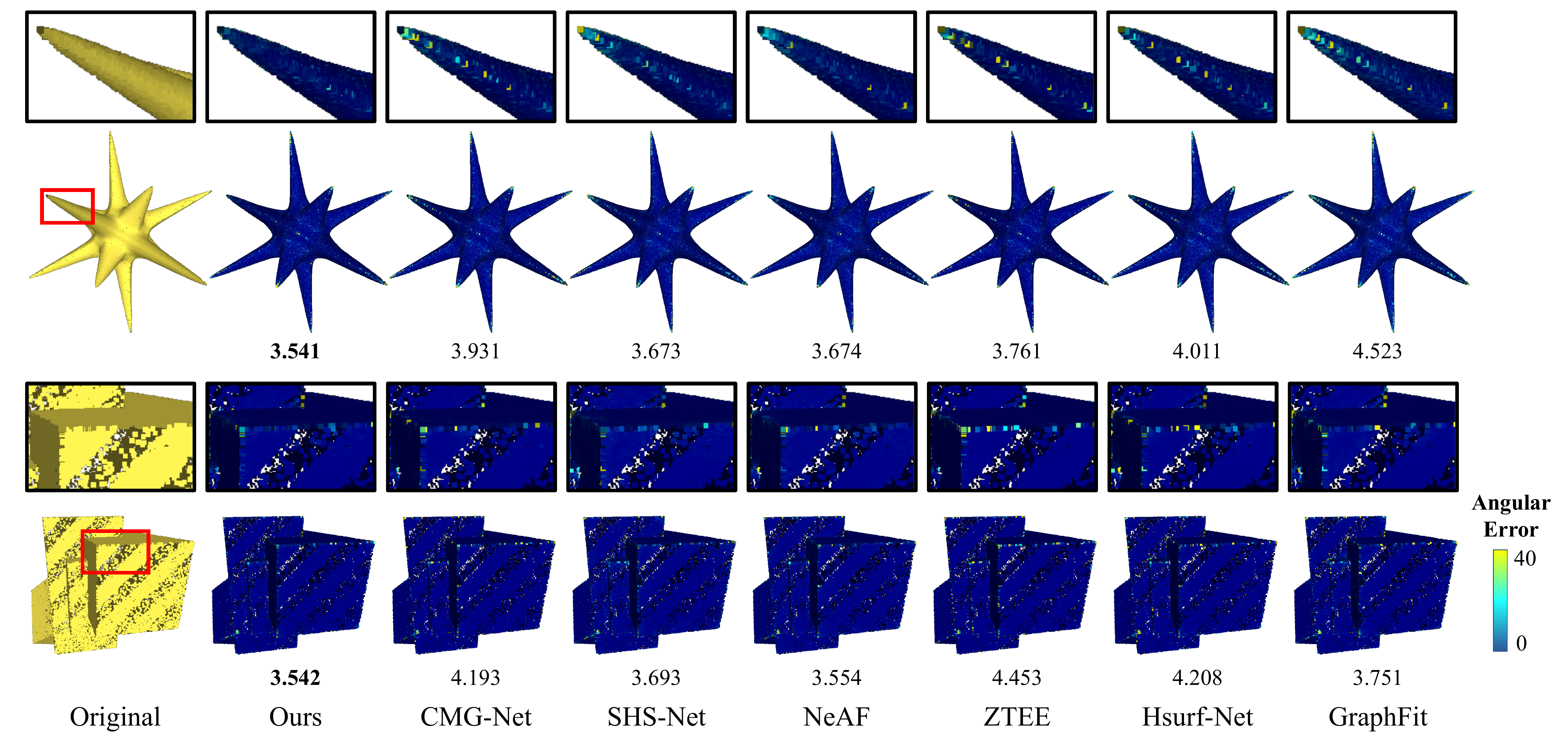}
  \caption{RMSE comparison on PCPNet test shapes with complex surface and uneven point distributions. The separate boxes present close-up regions, showing differences in local details.}
  \label{fig:pcpnet-closeup}
\end{figure*}

We first tested on PCPNet's test set, where the point clouds are sampled from mesh shapes. They contain ground-truth normals that are derived from mesh surfaces, forming an ideal benchmark for quantitative evaluation. The dataset consists of clean point clouds and their 5 variants, including 3 different noise levels (i.e., 0.12\%, 0.6\% and 1.2\%, corresponding to low, medium and high noise, respectively) and 2 types of point distributions (i.e., stripe and gradient). Following previous works, we sample a subset of 5,000 points from each shape using the dataset's provided indices for statistical evaluation. The RMSE results are demonstrated in Table~\ref{tab:pcpnet-rmse}, where our method achieves the minimum error on average. The PGP$\alpha$ results on each variant are shown in Fig.~\ref{fig:result-auc}, demonstrating our method's competitive performance. Moreover, Fig.~\ref{fig:pcpnet-closeup} shows close-up details regarding angular error on PCPNet's test shapes, showing our method's flexibility to estimate accurate normals on complex surfaces and uneven point distributions.


We further evaluated our method on another synthetic benchmark, the FamousShape dataset~\cite{li2023shsnet}, where the test shapes contain more complex geometric surfaces compared with PCPNet. This feature makes FamousShape a challenging benchmark for fitting-based normal estimation approaches. We compare our method against representative fitting-based baselines, including $n$-jet~\cite{cazals_2005_jet}, DeepFit~\cite{Shabat_deepfit_2020}, AdaFit~\cite{Zhu_adafit_2021}, GraphFit~\cite{li_graphfit_2022}, HSurf-Net~\cite{li_2022_hsurf}, NGLO~\cite{li_nglo_23} and SHS-Net~\cite{li2023shsnet}, and report the RMSE results in Table~\ref{tab:famousshape-fit}. Overall, our method achieves more accurate normal estimation results on diverse geometric structures, demonstrating its strong generalisation capability.

\begin{table}[t]  
\centering
\caption{RMSE on PCPNet's test set. \textbf{Bold} and \underline{underlined} values indicate the best and second-best results, respectively; the same convention is used in subsequent tables.}
\label{tab:pcpnet-rmse}
\setlength{\tabcolsep}{1mm}
\begin{tabular}{l|cccc|cc|c}
\toprule
\multirow{2}{*}{Method} & \multicolumn{4}{c|}{Noise}    & \multicolumn{2}{c|}{Density} &  \multirow{2}{*}{Avg.}  \\ \cmidrule{2-7}
                        & None  & Low    & Med.   & High  & Stripe       & Grad.      &                       \\ \midrule
PCA~\cite{Hoppe_1992}                     & 12.29 & 12.87  & 18.38 & \multicolumn{1}{c|}{27.52} & 13.66  & \multicolumn{1}{c|}{12.81}    & 16.25                 \\

$n$-Jet~\cite{cazals_2005_jet}                     & 12.35 & 12.84  & 18.33 & \multicolumn{1}{c|}{27.68} & 13.39  & \multicolumn{1}{c|}{13.13}    & 16.29                 \\

PCPNet~\cite{Guerrero_pcpnet_2018}                  & 9.64  & 11.51  & 18.27 & \multicolumn{1}{c|}{22.84} & 11.73  & \multicolumn{1}{c|}{13.46}    & 14.58                 \\

Nesti-Net~\cite{Benshabat_Nesti_2019}               & 7.06  & 10.24  & 17.77 & \multicolumn{1}{c|}{22.31} & 8.64   & \multicolumn{1}{c|}{8.95}     & 12.49                 \\

IterNet~\cite{Lenssen_DI_2020}                      & 6.72  & 9.95   & 17.18 & \multicolumn{1}{c|}{21.96} & 7.73   & \multicolumn{1}{c|}{7.51}     & 11.84                 \\

DeepFit~\cite{Shabat_deepfit_2020}                 & 6.51  & 9.21   & 16.73 & \multicolumn{1}{c|}{23.12} & 7.92   & \multicolumn{1}{c|}{7.31}     & 11.80                 \\

AdaFit~\cite{Zhu_adafit_2021}                  & 5.19  & 9.05   & 16.45 & \multicolumn{1}{c|}{21.94} & 6.01   & \multicolumn{1}{c|}{5.90}     & 10.76                 \\

GraphFit~\cite{li_graphfit_2022}                & 5.21  & 8.96   & 16.12 & \multicolumn{1}{c|}{21.71} & 6.30   & \multicolumn{1}{c|}{5.86}     & 10.69                 \\

HSurf-Net~\cite{li_2022_hsurf}                & 4.17  & 8.78   & 16.25 & \multicolumn{1}{c|}{21.61} & 4.98   & \multicolumn{1}{c|}{4.86}     & 10.11                 \\

NeAF~\cite{li2023neaf}                    & 4.20  & 9.25   & 16.35 & \multicolumn{1}{c|}{21.74} & 4.89   & \multicolumn{1}{c|}{4.88}     & 10.22                 \\

NGLO~\cite{li_nglo_23}                    & 4.06  & 8.70   & 16.12 & \multicolumn{1}{c|}{21.65} & 4.80   & \multicolumn{1}{c|}{4.56}     & 9.98                  \\

ZTEE~\cite{Du_2023_CVPR}               & \underline{3.85}  & 8.67   & \underline{16.11} & \multicolumn{1}{c|}{21.75} & \underline{4.78}   & \multicolumn{1}{c|}{4.63}     & 9.96           \\  

SHS-Net~\cite{li2023shsnet}                 & 3.95  & \underline{8.55}   & 16.13 & \multicolumn{1}{c|}{\underline{21.53}} & 4.91   & \multicolumn{1}{c|}{4.67}     & 9.96                  \\

CMG-Net~\cite{wu_cmgnet_24}                 & 3.86  & \textbf{8.45}   & \textbf{16.08} & \multicolumn{1}{c|}{21.89} & 4.85   & \multicolumn{1}{c|}{\textbf{4.45}}     & \underline{9.93}                  \\ \midrule 
Ours                    & \textbf{3.74}  & 8.76   & 16.25 & \multicolumn{1}{c|}{\textbf{21.52}} & \textbf{4.55}   & \multicolumn{1}{c|}{\underline{4.52}}     & \textbf{9.89}                  \\  

\bottomrule
\end{tabular}
\end{table}

\begin{table}[!ht]  
\centering
\caption{RMSE on FamousShape test set with other fitting-based normal estimation methods.}
\label{tab:famousshape-fit}
\setlength{\tabcolsep}{1mm}
\begin{tabular}{l|cccc|cc|c}
\toprule
\multirow{2}{*}{Method} & \multicolumn{4}{c|}{Noise}    & \multicolumn{2}{c|}{Density} &  \multirow{2}{*}{Avg.}  \\ \cmidrule{2-7}
                        & None  & Low    & Med.   & High  & Stripe       & Grad.      &                       \\ \midrule
$n$-Jet                & 20.11  & 20.57   & 31.34 & \multicolumn{1}{c|}{45.19} & 18.82   & \multicolumn{1}{c|}{18.69}     & 25.79                 \\                        
DeepFit                & 11.21  & 16.39   & 29.84 & \multicolumn{1}{c|}{39.95} & 11.84   & \multicolumn{1}{c|}{10.54}     & 19.96                 \\
AdaFit                    & 9.09  & 15.78   & 29.78 & \multicolumn{1}{c|}{38.74} & 8.52   & \multicolumn{1}{c|}{8.57}     & 18.41                  \\  
GraphFit                    & 8.91 & 15.73   & 29.37 & \multicolumn{1}{c|}{38.67} & 9.10   & \multicolumn{1}{c|}{8.62}     & 18.40                  \\  
HSurf-Net                    & 7.59  & 15.64   & 29.43 & \multicolumn{1}{c|}{\textbf{38.54}} & 7.63   & \multicolumn{1}{c|}{7.40}     & 17.70                  \\ 

NGLO                    & \underline{7.25}  & 15.60   & \underline{29.35} & \multicolumn{1}{c|}{38.74} & \underline{7.60}   & \multicolumn{1}{c|}{\textbf{7.20}}     & 17.62                  \\ 

SHS-Net                    & 7.41  & \textbf{15.34}   & \textbf{29.33} & \multicolumn{1}{c|}{38.56} & 7.74   & \multicolumn{1}{c|}{7.28}     & \underline{17.61}                  \\ 

\midrule
Ours                    & \textbf{7.22}  & \underline{15.40}   & 29.50 & \multicolumn{1}{c|}{\underline{38.55}} & \textbf{7.26}   & \multicolumn{1}{c|}{\underline{7.23}}     & \textbf{17.53}                  \\ 
\bottomrule
\end{tabular}
\end{table}

\subsection{Results on Real-world Benchmarks}

\begin{figure*}[t]
  \centering
  \includegraphics[width=\textwidth]{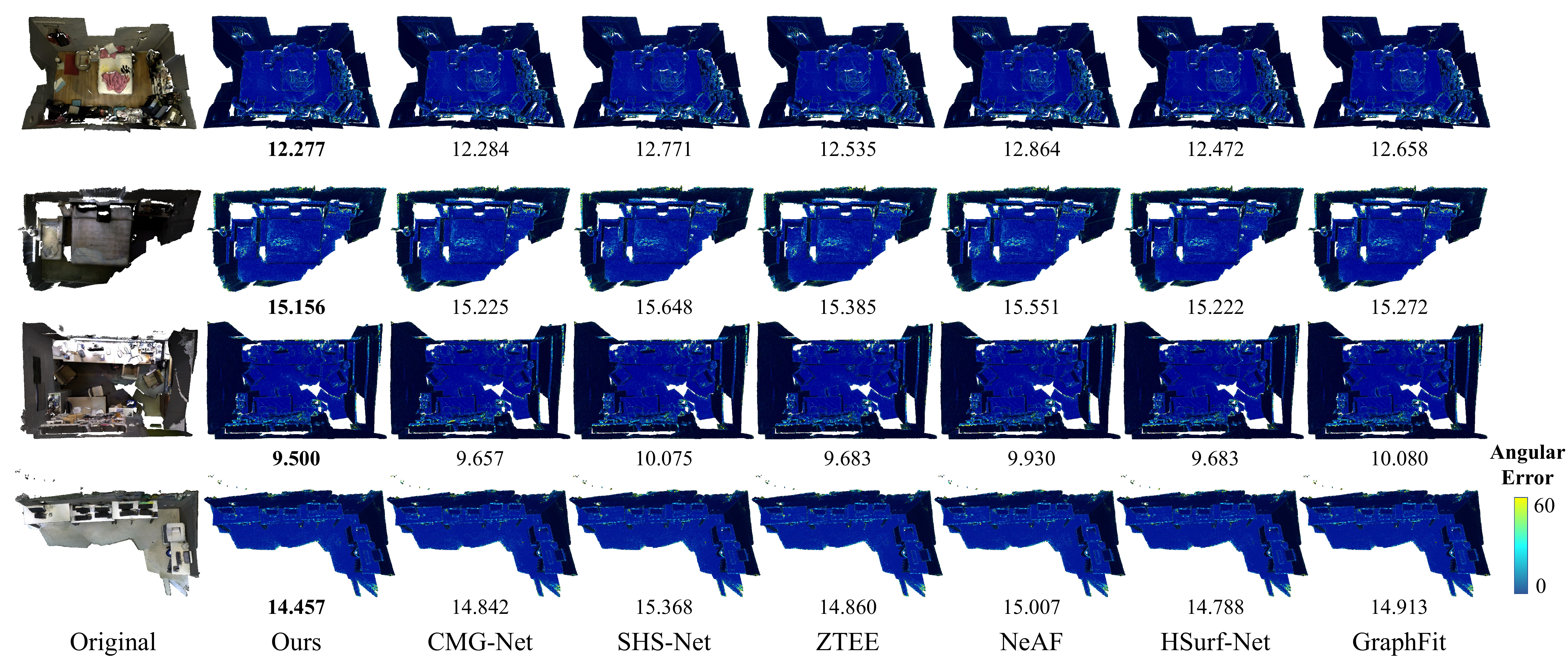}
      \caption{RMSE visualisation on scenes from SceneNN, which consists of complex real-world scanned indoor scenes.}
  \label{fig:result-sceneNN}
\end{figure*}

We tested on SceneNN~\cite{scenenn-3dv16}, a scanned dataset of indoor scenes covering diverse indoor environment types. Each scene has dense point samples (i.e., around 1 million) with ground-truth normals derived from reconstructed mesh surfaces.
We present statistical RMSE evaluation in Table~\ref{tab:sceneNN-rmse}, where our method has the smallest error on average, and the visualisation results in Fig.~\ref{fig:result-sceneNN} further demonstrate better feature preservation capability of our method compared with others.

\begin{figure*}[t]
  \centering
  \includegraphics[width=\textwidth]{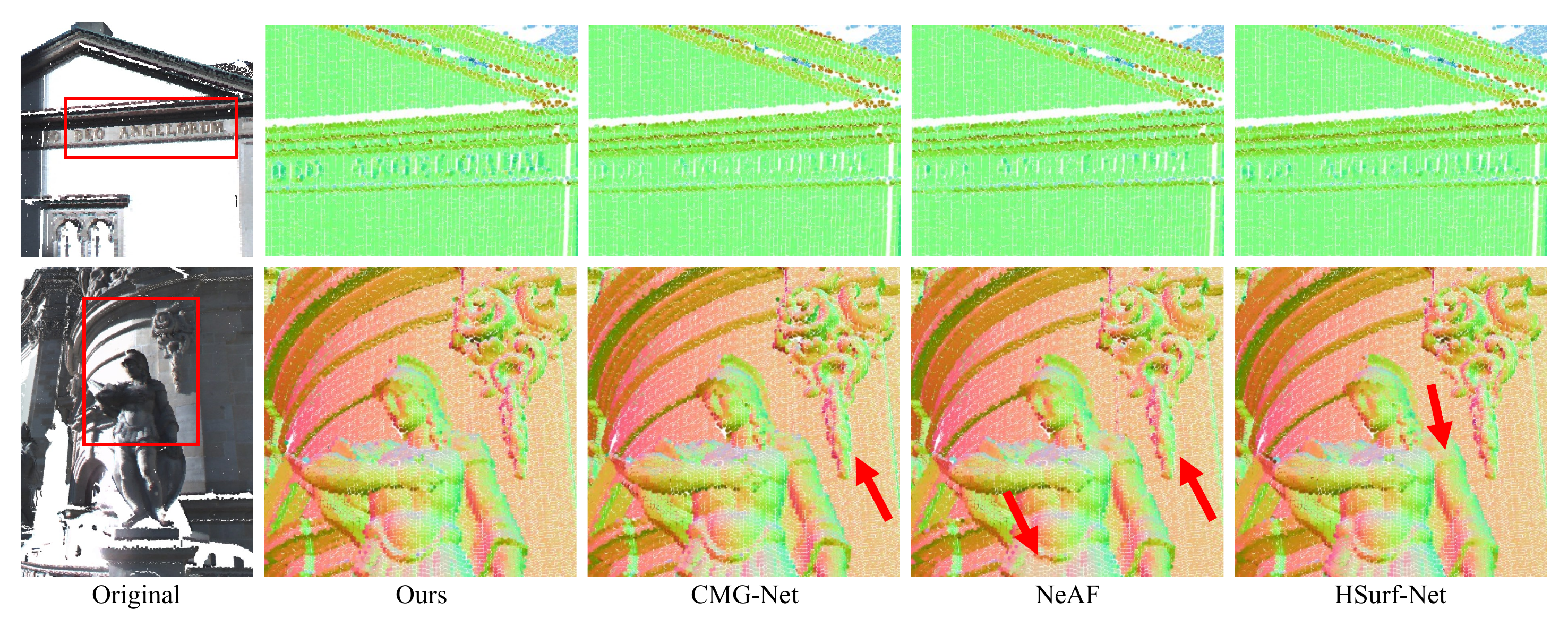}
  \caption{Normal estimation results on Semantic3D, where colours represent normal directions. Areas with blurred, inaccurate normals are marked by red arrows.}
  \label{fig:result-semantic3d}
\end{figure*}

We also evaluated on Semantic3D~\cite{semantics3d-hackel-17} which contains large-scale outdoor point clouds of street scenes captured via terrestrial laser scanning. We evaluate the visual results since ground-truth normals are not available in this dataset. The normal orientations are visualised in Fig.~\ref{fig:result-semantic3d} using the EAR software~\cite{EAR2013}, where our method achieves competitive results in terms of detail preservation.

\begin{table}[!ht]
\centering
\caption{Statistical RMSE evaluation on SceneNN dataset.}
\label{tab:sceneNN-rmse}
\setlength{\tabcolsep}{1mm}
\begin{tabular}{c|ccccccc}
\toprule
Method & Ours  & CMG-Net & SHS-Net & ZTEE  & NeAF  & HSurf-Net & GraphFit \\ \midrule
RMSE   & \textbf{13.30} & \underline{13.40}   & 13.91  & 13.53 & 13.78 & 13.46 & 13.69 \\
\bottomrule
\end{tabular}
\end{table}

\section{Discussions}

\subsection{Ablation Study}


To evaluate the contribution of each component in our framework, we perform a comprehensive ablation study covering the effects of major network modules, loss terms, and other training settings. The results are summarised in Table~\ref{tab:ablation}.

\subsubsection{Network Components}
We first assess the influence of removing the key modules from our network architecture. Specifically, we examine the impact of excluding the attention mechanism in our AHFF module and the Mamba blocks in our PSSM module. Both ablations lead to a decline in accuracy, highlighting these components' effectiveness in feature aggregation and surface modelling.

\subsubsection{Number of Mamba Blocks}
We then analyse the effect of varying the number of Mamba blocks in PSSM, where we test the configurations with 6 and 8 blocks. The results indicate that either increasing or decreasing the number of Mamba blocks degrades normal estimation accuracy.

\subsubsection{Alternative Loss Functions}
We investigate the effect of incorporating the $z$-direction transformation loss in ZTEE~\cite{Du_2023_CVPR}, denoted as $\mathcal{L}_\text{trans}$. To this end, we additionally introduce a QSTN network for $z$-axis alignment. However, this modification does not lead to a performance gain. We also remove the weighting loss $\mathcal{L}_\text{wt}$ and only train using $\mathcal{L}_\text{sin}$. Without the weighting mechanism based on tangent plane approximation, the network's performance is significantly degraded.

\subsubsection{Patch Size}
We test the impact of different patch sizes (600 and 800 points per patch, respectively) to analyse their influence on normal estimation. The results indicate that smaller patches provide less geometric context, while larger ones introduce redundancy to the network. Both settings result in degraded performance in terms of normal estimation accuracy.

\subsubsection{Alternative Network Components}
First, we compare our proposed AHFF module with the original attention-weighted module in SHS-Net~\cite{li2023shsnet}. Different from Eqs.~\eqref{eq:ahff-mean} and~\eqref{eq:attn-global}, in the original module, the global feature vector is calculated by
\begin{equation}
\mathbf{G}{s} = \Psi\left( \mathbf{v} \cdot \left( \text{MAX}\left\{\text{Softmax}_{N_s}(\mathbf{q}_d) \right\} \right)^\top \right),
\end{equation}
where the attention weights are aggregated via a \textit{max-pooling} operation. This strategy (denoted by Max-fusion in Table~\ref{tab:ablation}) only keeps the most dominant information within the encoding dimensions, which may overlook complementary geometric cues that coexist within the local point cloud regions. In contrast, our design (Eqs.~\eqref{eq:ahff-mean} and~\eqref{eq:attn-global}) keeps the \textit{mean} value and adds a learnable scale. Thus, our technique allows multiple correlated features to jointly contribute to the attention distribution, thereby leading to more stable feature representation.

Second, to evaluate different long-range dependency modelling strategies, we replace Mamba with Transformer which is implemented based on PCT~\cite{Guo_PCT_2021}. In PCT, the self-attention mechanism is defined as
\begin{equation}
    \mathbf{F}_1 = \text{AT}^1(\mathbf{F}_e),
\end{equation}
\begin{equation}
    \mathbf{F}_i = \text{AT}^i(\mathbf{F}_{i-1}), \quad i=2,3,4,
\end{equation}
\begin{equation}
    \mathbf{F}_o = \text{concat}(\mathbf{F}_1, \mathbf{F}_2, \mathbf{F}_3, \mathbf{F}_4) \cdot \mathbf{W}_o
\end{equation}
where $\mathbf{F}_e$ is the embedded point cloud feature, $\text{AT}^i$ indicates the $i$-th attention layer, and $\mathbf{W}_o$ represents the weights of a linear layer. For fair comparison, we set the encoding dimension to 128, which is consistent with our PSSM module. 

Third, we train our network model using the Chamfer Normal Distance (CND) strategy introduced in CMG-Net~\cite{wu_cmgnet_24}, where the normal of the nearest neighbour to each query point is adopted as the supervision normal during training. The experimental results in Table~\ref{tab:ablation} show that none of the three alternative strategies above outperform our design. In addition, Table~\ref{tab:cnd-vs} compares the CND metric between our model trained with and without this strategy, indicating that both variants achieve comparable performance without clear superiority.

\begin{table}[!t]
\centering
\caption{Ablation study on different training configurations and RMSE results on PCPNet's test set.}
\label{tab:ablation}
\setlength{\tabcolsep}{1mm}
\begin{tabular}{l|cccc|cc|c}
\toprule
\multirow{2}{*}{Config.} & \multicolumn{4}{c|}{Noise} & \multicolumn{2}{c|}{Density} & \multirow{2}{*}{Avg.} \\ \cmidrule{2-7}
                        & None  & Low  & Med.  & High & Stripe       & Grad.      &                       \\ \midrule
w/o Attn.   &  3.89     &  8.99     & 16.29     & 21.57     &   \underline{4.59}           &   \textbf{4.51}            &  9.97           \\
w/o Mamba           & 4.34  & 8.81 & \underline{16.16} & 21.57     &    5.29          &     5.11          &    10.21            \\ \midrule

6 Blocks                 &  \underline{3.83} & 8.77   & \underline{16.16} & \underline{21.50} &      4.64   &         4.58  &       \underline{9.91}         \\
8 Blocks                 &  3.95     & \underline{8.68}     &  \underline{16.16}     &  21.56    &   4.81       &   4.66       &    9.97    \\
\midrule

w/ $\mathcal{L}_\text{trans}$                & 4.17  & 8.78   & 16.25 & 21.61 & 4.98   & 4.86     & 10.11                 \\
w/o $\mathcal{L}_\text{wt}$               & 4.65  & 9.11   & 16.48 & 21.79 & 5.65   & 5.43     & 10.52                 \\
\midrule 

$N$ = 600                &  3.88     & 8.76      & 16.18      & 21.61      & 4.62              & 4.66              & 9.95                      \\
$N$ = 800                 & 4.23      & 8.73      & 16.34     & \textbf{21.36}      & 5.09              & 4.95              & 10.12                       \\ 
\midrule

Max-fusion                 & 3.89      & 8.89      & 16.26     & 21.51      & 4.78              & 4.59              & 9.99                       \\ 
Transformer                 & 4.21      & 8.85      & 16.19     & 21.61      & 5.05              & 4.97              & 10.15                       \\ 
w/ CND                & 4.02  & \textbf{8.65}   & \textbf{16.15} & 21.77 & 4.83   & 4.69     & 10.02                 \\
\midrule
Final                   &    \textbf{3.74}   &   8.76   &   16.25   &  21.52    &      \textbf{4.55}        &    \underline{4.52}           &        \textbf{9.89}               \\ \bottomrule
\end{tabular}
\end{table}

\begin{table}[t]  
\centering
\caption{CND evaluation on our model trained with and without this strategy on PCPNet's test set.}
\label{tab:cnd-vs}
\setlength{\tabcolsep}{1mm}
\begin{tabular}{l|cccc|cc|c}
\toprule
\multirow{2}{*}{Config.} & \multicolumn{4}{c|}{Noise}    & \multicolumn{2}{c|}{Density} &  \multirow{2}{*}{Avg.}  \\ \cmidrule{2-7}
                        & None  & Low    & Med.   & High  & Stripe       & Grad.      &                       \\ \midrule
w/ CND                & 4.02  & 8.36   & 12.75 & \multicolumn{1}{c|}{16.36} & 4.83   & \multicolumn{1}{c|}{4.69}     & 8.50                 \\
w/o CND                    & 3.74  & 8.47   & 13.29 & \multicolumn{1}{c|}{16.75} & 4.55   & \multicolumn{1}{c|}{4.52}     & 8.55                  \\  
\bottomrule
\end{tabular}
\end{table}

\subsection{Complexity Analysis}

We evaluate the computational complexity of our method and compare it with existing approaches. Specifically, we adopt the PyTorch-OpCounter tool (i.e., \texttt{thop}) to measure the number of parameters and floating point operations (FLOPs). We also consider a lightweight variant of our model, where the attention mechanism in AHFF is removed, and the number of Mamba blocks is set to 6. We denote this variant as \textit{Ours-L}. Then, we compare with the baseline method HSurf-Net~\cite{li_2022_hsurf} and other up-to-date methods, including ZTEE~\cite{Du_2023_CVPR}, SHS-Net~\cite{li2023shsnet} and CMG-Net~\cite{wu_cmgnet_24}. The FLOPs and number of parameters are outlined in Table~\ref{tab:complexity-analysis}, where we constantly set the patch size to $N = 700$ and the batch size to 256. It is noteworthy that compared with HSurf-Net, \textit{Ours-L} only slightly increases the parameter count and FLOPs, while achieving competitive results on PCPNet's test set, as shown in Table~\ref{tab:pcpnet-rmse-add}.

\begin{table}[t]
\centering
\caption{FLOPs and parameters of different methods. We skip the FLOPs comparison with SHS-Net as it requires an additional 1200-point input sample alongside each patch.}
\label{tab:complexity-analysis}
\setlength{\tabcolsep}{1mm}
\begin{tabular}{l|cc}
\toprule
Method    & FLOPs (G) & Params (M) \\
\midrule
HSurf-Net & \textbf{266.01}    & \textbf{2.16}      \\
ZTEE      & 1199.98   & 4.47       \\
SHS-Net   & -        & 3.27       \\
CMG-Net   & 714.69    & 2.70       \\
\midrule
Ours      & 444.77    & 6.61       \\
Ours-L   & \underline{278.01}    & \underline{2.42}      \\
\bottomrule
\end{tabular}
\end{table}

\begin{table}[t]  
\centering
\caption{RMSE on PCPNet's test set, where our lightweight variant achieves comparable performance.}
\label{tab:pcpnet-rmse-add}
\setlength{\tabcolsep}{1mm}
\begin{tabular}{l|cccc|cc|c}
\toprule
\multirow{2}{*}{Method} & \multicolumn{4}{c|}{Noise}    & \multicolumn{2}{c|}{Density} &  \multirow{2}{*}{Avg.}  \\ \cmidrule{2-7}
                        & None  & Low    & Med.   & High  & Stripe       & Grad.      &                       \\ \midrule

Ours                    & 3.74  & 8.76   & 16.25 & \multicolumn{1}{c|}{21.52} & 4.55   & \multicolumn{1}{c|}{4.52}     & 9.89                  \\  
Ours-L                    & 3.76  & 8.78   & 16.16 & \multicolumn{1}{c|}{21.60} & 4.76   & \multicolumn{1}{c|}{4.56}     & 9.94                  \\ 
\bottomrule
\end{tabular}
\end{table}

\begin{figure}[t]
  \centering
  \includegraphics[width=0.9\columnwidth]{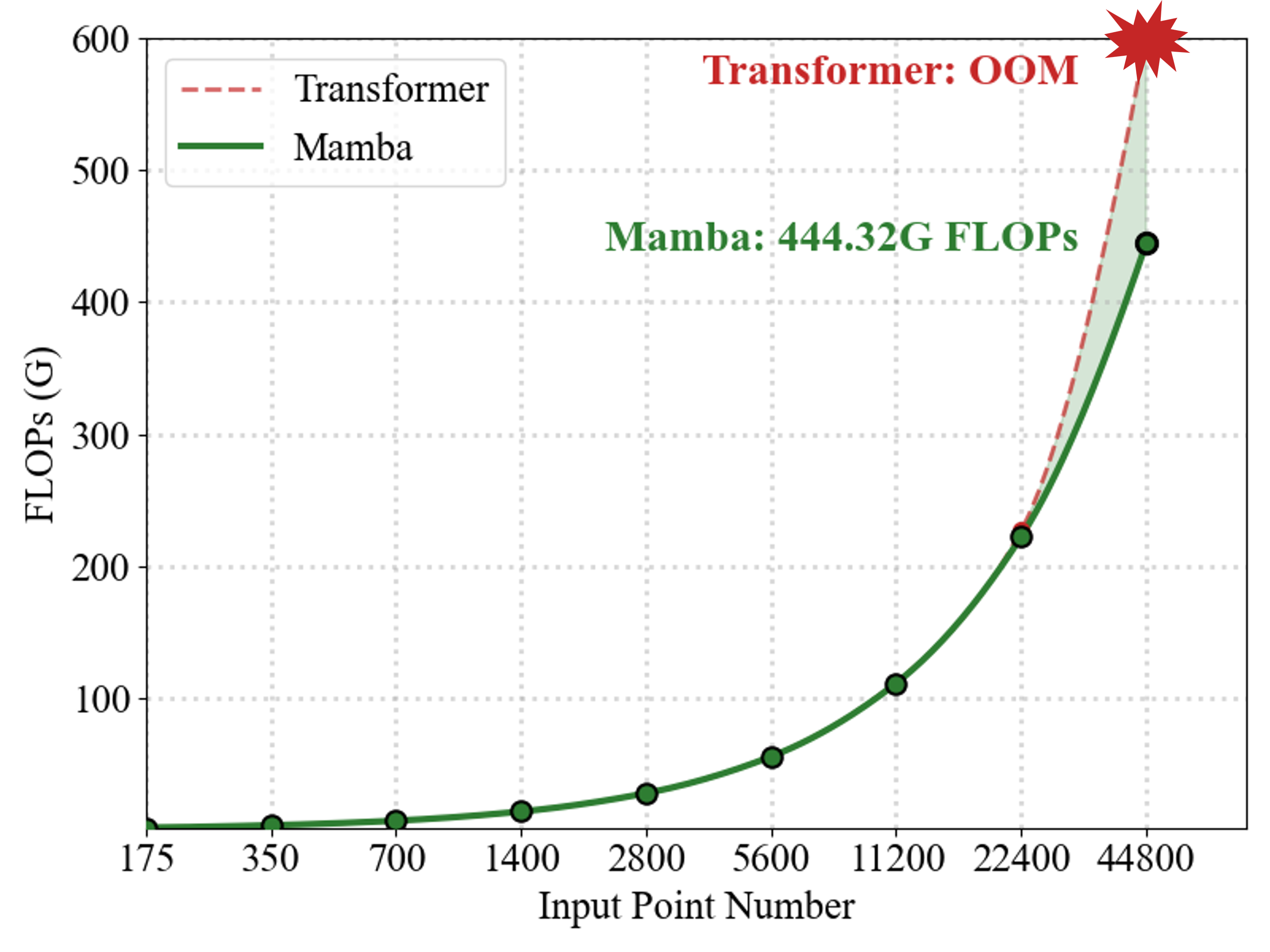}
  \caption{Stress test of Mamba-based and Transformer-based variants, where the batch size is constantly set to 1.}
  \label{fig:transformer_vs_mamba}
\end{figure}

Furthermore, we perform a memory stress test to illustrate the advantages of our Mamba-based framework. The test is performed by progressively increasing the input point numbers (i.e., token size $M$) on a 24 GB GPU, with the batch size fixed to 1. We observe that both variants perform comparably with smaller values regarding the input point number $M$. However, as $M$ increases, the Transformer variant runs out of memory at $M =$ 44,800, while the FLOPs of our Mamba variant are 444.32G, as shown in Fig.~\ref{fig:transformer_vs_mamba}.


\section{Applications}

\subsection{Surface Reconstruction}
\begin{figure}[t]
  \centering
  \includegraphics[width=\columnwidth]{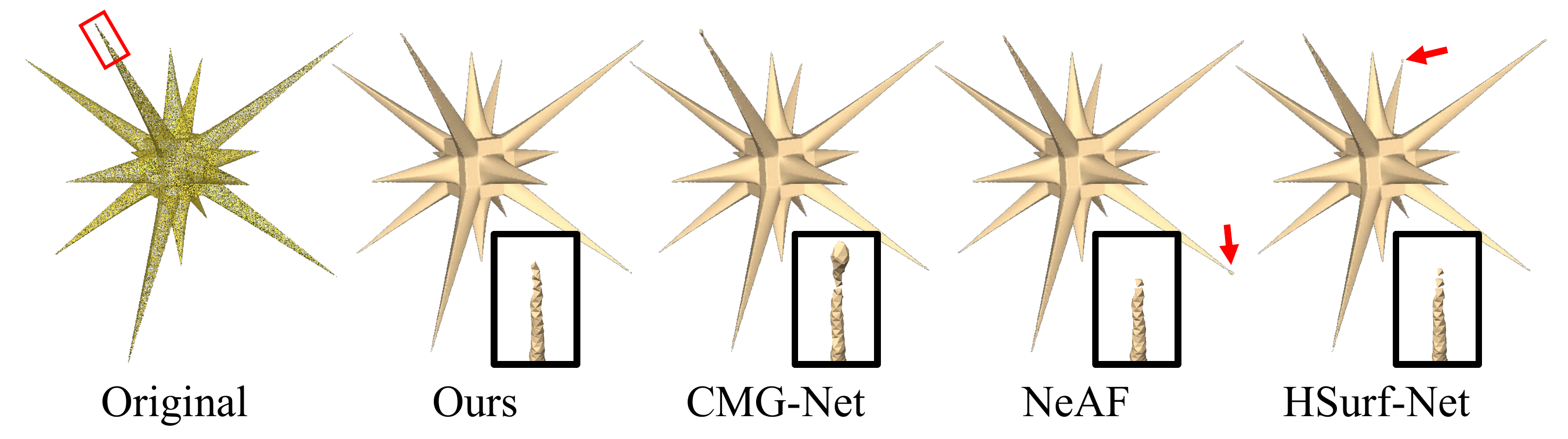}
  \caption{Mesh reconstruction results on a sharp star point cloud, where inaccurate areas are additionally marked by red arrows.}
  \label{fig:result-mesh}
\end{figure}

\begin{figure*}[t]
  \centering
  \includegraphics[width=0.9\textwidth]{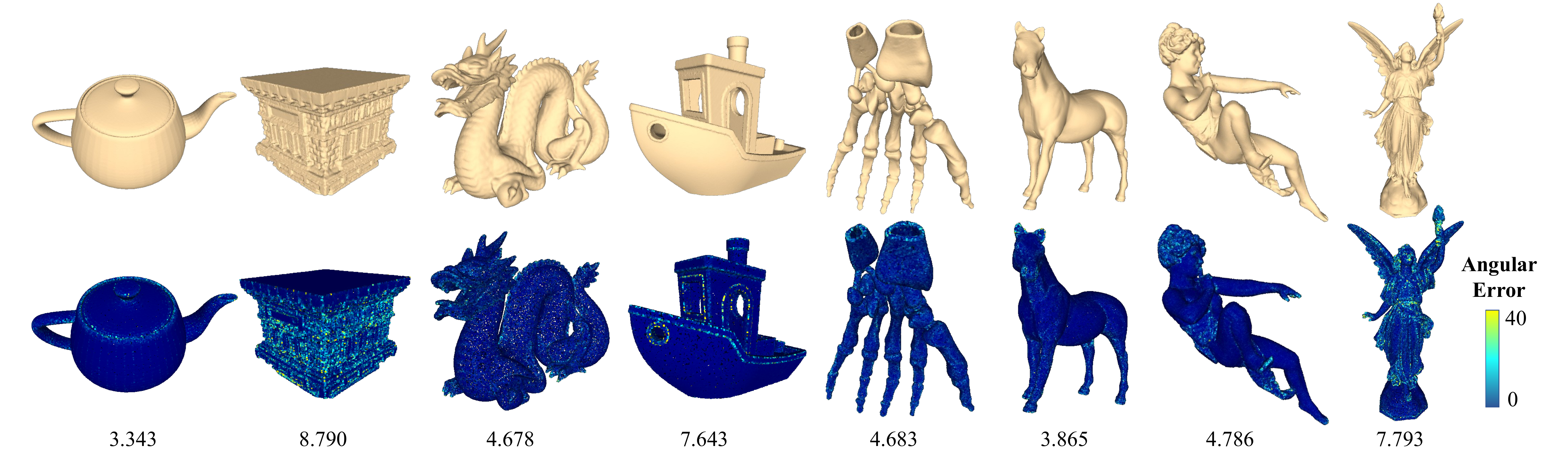}
  \caption{Surface reconstruction results and the corresponding RMSE on point clouds of our method on the FamousShape dataset.}
  \label{fig:result-famousshape-mesh}
\end{figure*}

Surface reconstruction serves as an important downstream application of normal estimation. Fig.~\ref{fig:result-mesh} shows the results of screened Poisson surface reconstruction~\cite{screened_poisson_2013} using the predicted normals on a challenging sharp star point cloud. Our proposed method provides more accurate normal prediction results, particularly on regions with high curvature, leading to improved surface reconstruction quality. In addition, Fig.~\ref{fig:result-famousshape-mesh} illustrates the reconstructed surfaces and the corresponding RMSE values on shapes from the FamousShape dataset, where the point clouds exhibit more complex geometric structures. Despite such increased complexity, our method still achieves accurate and visually consistent surface reconstruction results.

\subsection{Point Cloud Filtering}
\begin{figure*}[t]
  \centering
  \includegraphics[width=0.85\textwidth]{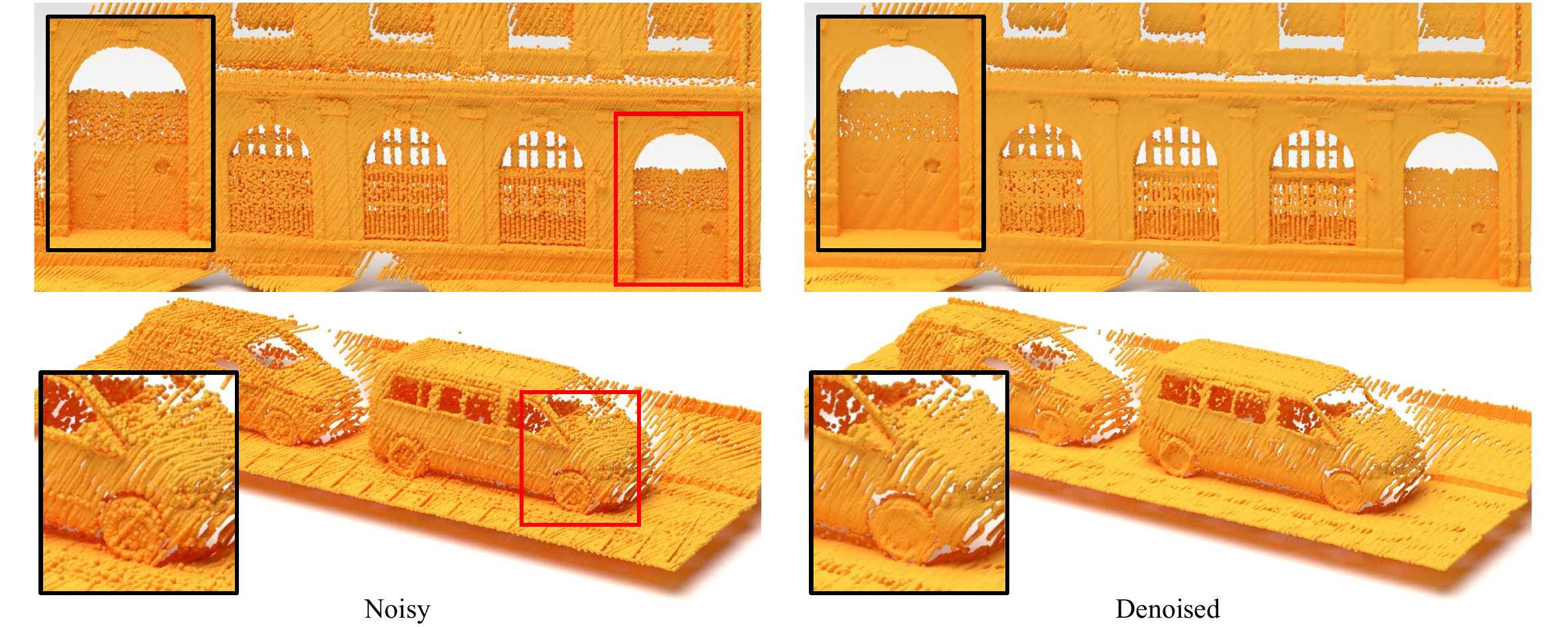}
  \caption{Point cloud filtering on Paris-rue-Madame via the low rank matrix approximation method, using our normal prediction results as input.}
  \label{fig:result-paris-denoise}
\end{figure*}

Normals also play a critical role in point cloud filtering. We adopt the Paris Rue Madame~\cite{serna-paris-2014} dataset, a 3D mobile laser scanning dataset that captures street scenes and is severely contaminated by noise. To perform filtering, we use the low rank matrix approximation method~\cite{Lu_lowrank_2020} using our predicted normals as input. The filtering results are presented in Fig.~\ref{fig:result-paris-denoise}, where the point cloud surfaces are significantly smoothed.

\section{Conclusion}
In this paper, we introduced MambaH-Fit, a novel SSM-based framework for point cloud normal estimation. We first designed AHFF, an attention-driven hierarchical feature fusion module that adaptively extracts global features from larger-scale point cloud neighbourhoods and fuses them into smaller ones, thereby facilitating the learning of detailed geometric structures. We also proposed PSSM, a patch-wise SSM tailored for high-quality hyper-surface fitting, enabling effective modelling of fine-grained local geometric structures. Comprehensive experiments demonstrate that our method achieves competitive results across diverse datasets, outperforming existing methods in terms of accuracy, robustness, and generalisation capability to large-scale real-world scanned point clouds.



\bibliographystyle{IEEEtran}
\bibliography{bibliography}

\end{document}